\documentclass[10pt,onecolumn,letterpaper]{article}

\usepackage{iccv}
\usepackage{times}
\usepackage{epsfig}
\usepackage{graphicx}
\usepackage{amsmath}
\usepackage{amssymb}
\usepackage{graphicx}
\usepackage{amsmath}
\usepackage{amssymb}
\usepackage{booktabs}
\usepackage{cite}
\usepackage{booktabs}
\usepackage{multirow}
\usepackage[table,xcdraw]{xcolor}

\usepackage[pagebackref=true,breaklinks=true,letterpaper=true,colorlinks,bookmarks=false]{hyperref}

\iccvfinalcopy 

\begin{document}


\title{Evaluating SAM2’s Role in Camouflaged Object Detection: From SAM to SAM2}

\author{Lv Tang  \quad Bo Li \\
{vivo Mobile Communication Co., Ltd, Shanghai, China} \\
{\tt\small luckybird1994@gmail.com}, 
{\tt\small libra@vivo.com}}
\maketitle

\begin{abstract}
The Segment Anything Model (SAM), introduced by Meta AI Research as a generic object segmentation model, quickly garnered widespread attention and significantly influenced the academic community. To extend its application to video, Meta further develops Segment Anything Model 2 (SAM2), a unified model capable of both video and image segmentation. SAM2 shows notable improvements over its predecessor in terms of applicable domains, promptable segmentation accuracy, and running speed. However, this report reveals a decline in SAM2’s ability to perceive different objects in images without prompts in its auto mode, compared to SAM. Specifically, we employ the challenging task of camouflaged object detection to assess this performance decrease, hoping to inspire further exploration of the SAM model family by researchers. The results of this paper are provided in \url{https://github.com/luckybird1994/SAMCOD}.
\end{abstract}

\section{Introduction}
In recent years, large language models (LLMs)~\cite{chowdhery2023palm,touvron2023llama,zhang2022opt} have sparked a revolution in natural language processing (NLP). These foundational models exhibit remarkable transfer capabilities, extending far beyond their initial training objectives. LLMs showcase robust generalization abilities and excel in a multitude of open-world language tasks, including language comprehension, generation, interaction, and reasoning. Inspired by the success of LLMs, vision foundational models such as CLIP~\cite{DBLP:conf/icml/RadfordKHRGASAM21}, DINOv2~\cite{DBLP:journals/corr/abs-2304-07193}, BLIP~\cite{DBLP:conf/icml/0001LXH22}, and SAM~\cite{kirillov2023segment} have also emerged. The introduction of these foundational models continues to drive researchers’ exploration in the field of computer vision.

Among these foundational models, SAM stands out as an influential model in the domain of image segmentation. Upon its introduction, it quickly receives widespread attention and is utilized across various fields. To further extend SAM’s applicability, SAM2~\cite{ravi2024sam2} is introduced, designed to handle both image and video segmentation tasks within a unified architecture. Compared to SAM, SAM2 has significantly improved in terms of promptable segmentation accuracy and running speed, further enhancing the capabilities of the SAM model family.

In this technical report, we use the task of camouflaged object detection (COD) as a case study to analyze the progress and compromises involved in transitioning from SAM to SAM2. We observe the following two points: 1) When the SAM2 model is provided with prompts for segmentation, its performance shows substantial improvement compared to SAM. 2) However, when operating in auto mode, where SAM2 and SAM are both tasked with perceiving all objects in an image without prompts, the performance of SAM2 significantly deteriorates. We hope that our findings will further encourage researchers to explore both SAM and SAM2.

\section{Experiments}\label{sec4}
\subsection{Datasets and Metrics}
To validate the performance of SAM and SAM2, we evaluate its performance on three benchmark datasets, containing CAMO~\cite{DBLP:journals/cviu/LeNNTS19}, COD10K~\cite{DBLP:conf/cvpr/FanJSCS020}, NC4K~\cite{DBLP:conf/cvpr/Lv0DLLBF21} and MoCA-Mask~\cite{DBLP:conf/cvpr/ChengXFZHDG22}. We use six widely used metrics: structure-measure ($S_\alpha$)~\cite{DBLP:conf/iccv/FanCLLB17}, mean E-measure ($E_\phi$)~\cite{21Fan_HybridLoss}, F-measure ($F_\beta$), weighted F-measure ($F^w_\beta$), max F-measure ($F^{max}_\beta$)~\cite{DBLP:conf/cvpr/MargolinZT14}, and mean absolute error (MAE).

\subsection{Progress}

\begin{table}[!htbp]
\centering
\caption{Comparison results between SAM2 and two state-of-the-art VCOD methods}
\begin{tabular}{@{}c|ccc@{}}
\toprule
                          & \multicolumn{3}{c}{MoCA-Mask}                                                                 \\ \cmidrule(l){2-4} 
\multirow{-2}{*}{Methods} & $F_\beta^w$                   & $S_\alpha$                    & MAE                           \\ \midrule
SLTNet (CVPR2022)         & \cellcolor[HTML]{FFFFFF}0.357 & \cellcolor[HTML]{FFFFFF}0.656 & \cellcolor[HTML]{FFFFFF}0.021 \\
TSP-SAM(CVPR2024)         & \cellcolor[HTML]{FFFFFF}0.444 & \cellcolor[HTML]{FFFFFF}0.689 & \cellcolor[HTML]{FFFFFF}0.008 \\
SAM2                      & \cellcolor[HTML]{FFFFFF}0.691 & \cellcolor[HTML]{FFFFFF}0.804 & \cellcolor[HTML]{FFFFFF}0.004 \\ \bottomrule
\end{tabular}
\label{sam2vcod}
\end{table}
\textbf{Video Task.} One of the biggest advancements of SAM2 over SAM is its application in video tasks. Therefore, we first test the potential of SAM2 in the VCOD task, selecting the MoCA-Mask as the evaluation dataset. Specifically, for the first frame of each video sequence, we randomly select three prompt points based on its corresponding ground truth to identify the target object for segmentation. The video sequence is then input into the SAM2, and the segmentation results are obtained as shown in Table. \ref{sam2vcod}. SAM2 significantly surpasses state-of-the-art VCOD methods SLTNet~\cite{DBLP:conf/cvpr/ChengXFZHDG22} and TSP-SAM~\cite{hui2024endow}.

\begin{table}[!htbp]
\centering
\caption{Promptable segmentation performance of SAM and SAM2.}
\scalebox{0.9}{
\begin{tabular}{@{}c|c|ccc|ccc|ccc@{}}
\toprule
 &
   &
  \multicolumn{3}{c|}{CAMO (250 Images)} &
  \multicolumn{3}{c|}{COD10K (2026 Images)} &
  \multicolumn{3}{c}{NC4K (4121 Images)} \\ \cmidrule(l){3-11} 
\multirow{-2}{*}{Methods} &
  \multirow{-2}{*}{Setting} &
  $F_{\beta}^{w}$ &
  $S_{\alpha}$ &
  MAE &
  $F_{\beta}^{w}$ &
  $S_{\alpha}$ &
  MAE &
  $F_{\beta}^{w}$ &
  $S_{\alpha}$ &
  MAE \\ \midrule
FSPNet (CVPR2023) &
  F &
  {\color[HTML]{333333} 0.799} &
  {\color[HTML]{333333} 0.856} &
  {\color[HTML]{333333} 0.050} &
  {\color[HTML]{333333} 0.735} &
  {\color[HTML]{333333} 0.851} &
  {\color[HTML]{333333} 0.026} &
  {\color[HTML]{333333} 0.816} &
  {\color[HTML]{333333} 0.879} &
  {\color[HTML]{333333} 0.035} \\
NCHIT (CVIU2022) &
  F &
  {\color[HTML]{333333} 0.652} &
  {\color[HTML]{333333} 0.784} &
  {\color[HTML]{333333} 0.088} &
  {\color[HTML]{333333} 0.591} &
  {\color[HTML]{333333} 0.792} &
  {\color[HTML]{333333} 0.049} &
  {\color[HTML]{333333} 0.710} &
  {\color[HTML]{333333} 0.830} &
  {\color[HTML]{333333} 0.058} \\
ERRNet (PR2022) &
  F &
  {\color[HTML]{333333} 0.679} &
  {\color[HTML]{333333} 0.779} &
  {\color[HTML]{333333} 0.085} &
  {\color[HTML]{333333} 0.630} &
  {\color[HTML]{333333} 0.786} &
  {\color[HTML]{333333} 0.043} &
  {\color[HTML]{333333} 0.737} &
  {\color[HTML]{333333} 0.827} &
  {\color[HTML]{333333} 0.054} \\ \midrule

SAM (Shikra+SAM) &
  ZS &
  {\color[HTML]{333333} 0.521} &
  {\color[HTML]{333333} 0.652} &
  {\color[HTML]{333333} 0.128} &
  {\color[HTML]{333333} 0.482} &
  {\color[HTML]{333333} 0.657} &
  {\color[HTML]{333333} 0.111} &
  {\color[HTML]{333333} 0.570} &
  {\color[HTML]{333333} 0.689} &
  {\color[HTML]{333333} 0.120} \\ 
SAM2 (Shikra+SAM2) &
  ZS &
  {\color[HTML]{333333} 0.620} &
  {\color[HTML]{333333} 0.716} &
  {\color[HTML]{333333} 0.113} &
  {\color[HTML]{333333} 0.565} &
  {\color[HTML]{333333} 0.708} &
  {\color[HTML]{333333} 0.101} &
  {\color[HTML]{333333} 0.672} &
  {\color[HTML]{333333} 0.760} &
  {\color[HTML]{333333} 0.092} \\ \midrule

SAM (LLAVA+SAM) &
  ZS &
  {\color[HTML]{333333} 0.520} &
  {\color[HTML]{333333} 0.647} &
  {\color[HTML]{333333} 0.141} &
  {\color[HTML]{333333} 0.552} &
  {\color[HTML]{333333} 0.696} &
  {\color[HTML]{333333} 0.094} &
  {\color[HTML]{333333} 0.591} &
  {\color[HTML]{333333} 0.699} &
  {\color[HTML]{333333} 0.115} \\ 
SAM2 (LLAVA+SAM2) &
  ZS &
  {\color[HTML]{333333} 0.633} &
  {\color[HTML]{333333} 0.722} &
  {\color[HTML]{333333} 0.114} &
  {\color[HTML]{333333} 0.640} &
  {\color[HTML]{333333} 0.754} &
  {\color[HTML]{333333} 0.078} &
  {\color[HTML]{333333} 0.700} &
  {\color[HTML]{333333} 0.776} &
  {\color[HTML]{333333} 0.085} \\ \bottomrule
\end{tabular}}
\label{mllmcod}
\end{table}

\textbf{Promptable Segmentation.} We further evaluate the performance of the SAM and SAM2 on promptable segmentation. Inspired by the recent study~\cite{tang2024chain}, we use MLLMs, Shikra~\cite{DBLP:journals/corr/abs-2306-15195} and LLaVA~\cite{DBLP:journals/corr/abs-2304-08485} to generate coordinates for camouflaged objects in images. These coordinates are then input into both SAM and SAM2 to produce the corresponding mask results. The performance is shown in Table. \ref{mllmcod}. it is evident that SAM2’s performance significantly exceeds that of SAM. Moreover, through this approach, the performance of zero-shot CoD methods has the potential to match or surpass fully supervised methods. These findings highlight the advancements made by SAM2.

\clearpage

\subsection{Compromise}
\begin{table}[!htbp]
\centering
\caption{Comparison results of SAM and SAM2 in Auto Mode.}
\scalebox{0.7}{
\begin{tabular}{@{}cccccccccccccccccccc@{}}
\toprule
\multicolumn{1}{c|}{}                                             & \multicolumn{1}{c|}{}                                                  & \multicolumn{6}{c|}{\textbf{CAMO-Test (250 images)}}                                                                                                                                                                                 & \multicolumn{6}{c|}{COD10K-Test (2026 images)}                                                                                                                                                               & \multicolumn{6}{c}{\textbf{NC4K (4121 images)}}                                                                                                                                         \\ \cmidrule(l){3-20} 
\multicolumn{1}{c|}{\multirow{-2}{*}{Methods}}                    & \multicolumn{1}{c|}{\multirow{-2}{*}{\textbf{Pub.}}}                   & $S_\alpha$                                           & $E_\phi$                     & $F^w_\beta$                  & $F_\beta$                    & $F^{max}_\beta$              & \multicolumn{1}{c|}{MAE}                          & $S_\alpha$                   & $E_\phi$                     & $F^w_\beta$                  & $F_\beta$                    & $F^{max}_\beta$              & \multicolumn{1}{c|}{MAE}                          & $S_\alpha$                   & $E_\phi$                     & $F^w_\beta$                  & $F_\beta$                    & $F^{max}_\beta$              & MAE                          \\ \midrule
\multicolumn{20}{c}{CNN-Based Models}                                                                                                                                                                                                                                                                                                                                                                                                                                                                                                                                                                                                                                                                                                                                                      \\ \midrule
\multicolumn{1}{c|}{SINet}                                        & \multicolumn{1}{c|}{CVPR'20}                                           & 0.751                                                & 0.771                        & 0.606                        & 0.675                        & 0.706                        & \multicolumn{1}{c|}{0.100}                        & 0.771                        & 0.806                        & 0.551                        & 0.634                        & 0.676                        & \multicolumn{1}{c|}{0.051}                        & 0.808                        & 0.871                        & 0.723                        & 0.769                        & 0.775                        & 0.058                        \\
\multicolumn{1}{c|}{C2FNet}                                       & \multicolumn{1}{c|}{IJCAI'21}                                          & 0.796                                                & 0.854                        & 0.719                        & 0.762                        & 0.771                        & \multicolumn{1}{c|}{0.080}                        & 0.812                        & 0.890                        & 0.686                        & 0.723                        & 0.743                        & \multicolumn{1}{c|}{0.036}                        & 0.838                        & 0.897                        & 0.762                        & 0.795                        & 0.810                        & 0.049                        \\
\multicolumn{1}{c|}{LSR}                                          & \multicolumn{1}{c|}{CVPR'21}                                           & 0.787                                                & 0.838                        & 0.696                        & 0.744                        & 0.753                        & \multicolumn{1}{c|}{0.080}                        & 0.804                        & 0.880                        & 0.673                        & 0.715                        & 0.732                        & \multicolumn{1}{c|}{0.037}                        & 0.840                        & 0.895                        & 0.766                        & 0.804                        & 0.815                        & 0.048                        \\
\multicolumn{1}{c|}{PFNet}                                        & \multicolumn{1}{c|}{CVPR'21}                                           & 0.782                                                & 0.841                        & 0.695                        & 0.746                        & 0.758                        & \multicolumn{1}{c|}{0.085}                        & 0.800                        & 0.877                        & 0.660                        & 0.701                        & 0.725                        & \multicolumn{1}{c|}{0.040}                        & 0.829                        & 0.887                        & 0.745                        & 0.784                        & 0.799                        & 0.053                        \\
\multicolumn{1}{c|}{MGL}                                          & \multicolumn{1}{c|}{CVPR'21}                                           & 0.775                                                & 0.812                        & 0.673                        & 0.726                        & 0.740                        & \multicolumn{1}{c|}{0.088}                        & 0.814                        & 0.851                        & 0.666                        & 0.710                        & 0.738                        & \multicolumn{1}{c|}{0.035}                        & 0.833                        & 0.867                        & 0.739                        & 0.782                        & 0.800                        & 0.053                        \\
\multicolumn{1}{c|}{JCOD}                                         & \multicolumn{1}{c|}{CVPR'21}                                           & 0.800                                                & 0.859                        & 0.728                        & 0.772                        & 0.779                        & \multicolumn{1}{c|}{0.073}                        & 0.809                        & 0.884                        & 0.684                        & 0.721                        & 0.738                        & \multicolumn{1}{c|}{0.035}                        & 0.842                        & 0.898                        & 0.771                        & 0.806                        & 0.816                        & 0.047                        \\
\multicolumn{1}{c|}{TANet}                                        & \multicolumn{1}{c|}{AAAI'21}                                           & 0.781                                                & 0.847                        & 0.678                        & -                            & -                            & \multicolumn{1}{c|}{0.087}                        & 0.793                        & 0.848                        & 0.635                        & -                            & -                            & \multicolumn{1}{c|}{0.043}                        & -                            & -                            & -                            & -                            & -                            & -                            \\
\multicolumn{1}{c|}{BGNet}                                        & \multicolumn{1}{c|}{IJCAI'22}                                          & 0.813                                                & 0.870                        & 0.749                        & 0.789                        & 0.799                        & \multicolumn{1}{c|}{0.073}                        & 0.831                        & {\color[HTML]{330001} 0.901} & 0.722                        & 0.753                        & 0.774                        & \multicolumn{1}{c|}{0.033}                        & 0.851                        & 0.907                        & {\color[HTML]{FE0000} 0.788} & 0.820                        & {\color[HTML]{FE0000} 0.833} & 0.044                        \\
\multicolumn{1}{c|}{FDCOD}                                        & \multicolumn{1}{c|}{CVPR'22}                                           & 0.828                                                & 0.883                        & 0.748                        & 0.781                        & 0.804                        & \multicolumn{1}{c|}{0.068}                        & 0.832                        & {\color[HTML]{FE0000} 0.907} & 0.706                        & 0.733                        & 0.776                        & \multicolumn{1}{c|}{0.033}                        & 0.834                        & 0.893                        & 0.750                        & {\color[HTML]{FE0000} 0.784} & 0.804                        & 0.051                        \\
\multicolumn{1}{c|}{SegMaR}                                       & \multicolumn{1}{c|}{CVPR'22}                                           & 0.815                                                & 0.874                        & 0.753                        & 0.795                        & 0.803                        & \multicolumn{1}{c|}{0.071}                        & 0.833                        & 0.899                        & 0.724                        & 0.757                        & 0.774                        & \multicolumn{1}{c|}{0.034}                        & 0.841                        & 0.896                        & 0.781                        & 0.821                        & 0.826                        & 0.046                        \\
\multicolumn{1}{c|}{ZoomNet}                                      & \multicolumn{1}{c|}{CVPR'22}                                           & 0.820                                                & 0.877                        & 0.752                        & 0.794                        & 0.805                        & \multicolumn{1}{c|}{0.066}                        & {\color[HTML]{FE0000} 0.838} & 0.888                        & {\color[HTML]{FE0000} 0.729} & {\color[HTML]{FE0000} 0.766} & {\color[HTML]{FE0000} 0.780} & \multicolumn{1}{c|}{{\color[HTML]{FE0000} 0.029}} & 0.853                        & 0.896                        & 0.784                        & 0.818                        & 0.828                        & 0.043                        \\
\multicolumn{1}{c|}{BSANet}                                       & \multicolumn{1}{c|}{AAAI'22}                                           & 0.794                                                & 0.851                        & 0.717                        & 0.763                        & 0.770                        & \multicolumn{1}{c|}{0.079}                        & 0.817                        & 0.891                        & 0.699                        & 0.738                        & 0.753                        & \multicolumn{1}{c|}{0.034}                        & 0.841                        & 0.897                        & 0.771                        & 0.808                        & 0.817                        & 0.048                        \\ \midrule
\multicolumn{1}{c|}{SINetV2}                                      & \multicolumn{1}{c|}{PAMI'22}                                           & 0.822                                                & 0.882                        & 0.743                        & 0.782                        & 0.801                        & \multicolumn{1}{c|}{0.070}                        & 0.815                        & 0.887                        & 0.680                        & 0.718                        & 0.752                        & \multicolumn{1}{c|}{0.037}                        & 0.847                        & 0.903                        & 0.770                        & 0.805                        & 0.823                        & 0.048                        \\
\multicolumn{1}{c|}{FAPNet}                                       & \multicolumn{1}{c|}{TIP'22}                                            & 0.815                                                & 0.865                        & 0.734                        & 0.776                        & 0.792                        & \multicolumn{1}{c|}{0.076}                        & 0.822                        & 0.887                        & 0.694                        & 0.731                        & 0.758                        & \multicolumn{1}{c|}{0.036}                        & 0.851                        & 0.899                        & 0.775                        & 0.810                        & 0.825                        & 0.046                        \\
\multicolumn{1}{c|}{C2FNet-ext}                                   & \multicolumn{1}{c|}{TCSVT'22}                                          & 0.799                                                & 0.859                        & 0.730                        & 0.770                        & 0.779                        & \multicolumn{1}{c|}{0.077}                        & 0.811                        & 0.887                        & 0.691                        & 0.725                        & 0.742                        & \multicolumn{1}{c|}{0.036}                        & 0.840                        & 0.896                        & 0.770                        & 0.802                        & 0.814                        & 0.048                        \\
\multicolumn{1}{c|}{DGNet}                                        & \multicolumn{1}{c|}{MIR'22}                                            & 0.839                                                & {\color[HTML]{FE0000} 0.900} & 0.768                        & {\color[HTML]{FE0000} 0.806} & 0.822                        & \multicolumn{1}{c|}{{\color[HTML]{FE0000} 0.057}} & 0.822                        & 0.896                        & 0.693                        & 0.728                        & 0.759                        & \multicolumn{1}{c|}{0.033}                        & {\color[HTML]{FE0000} 0.854} & 0.909                        & 0.783                        & 0.813                        & 0.830                        & {\color[HTML]{FE0000} 0.043} \\
\multicolumn{1}{c|}{CubeNet}                                      & \multicolumn{1}{c|}{PR'22}                                             & 0.788                                                & 0.838                        & 0.682                        & 0.732                        & 0.750                        & \multicolumn{1}{c|}{0.085}                        & 0.795                        & 0.865                        & 0.643                        & 0.692                        & 0.715                        & \multicolumn{1}{c|}{0.041}                        & -                            & -                            & -                            & -                            & -                            & -                            \\
\multicolumn{1}{c|}{ERRNet}                                       & \multicolumn{1}{c|}{PR'22}                                             & 0.779                                                & 0.842                        & 0.679                        & 0.729                        & 0.742                        & \multicolumn{1}{c|}{0.085}                        & 0.786                        & 0.867                        & 0.630                        & 0.675                        & 0.702                        & \multicolumn{1}{c|}{0.043}                        & 0.827                        & 0.887                        & 0.737                        & 0.778                        & 0.794                        & 0.054                        \\ \midrule
\multicolumn{20}{c}{Transformer-Based Models}                                                                                                                                                                                                                                                                                                                                                                                                                                                                                                                                                                                                                                                                                                                                              \\ \midrule
\multicolumn{1}{c|}{VST}                                          & \multicolumn{1}{c|}{ICCV'21}                                           & 0.807                                                & 0.848                        & 0.713                        & 0.758                        & 0.777                        & \multicolumn{1}{c|}{0.081}                        & 0.820                        & 0.879                        & 0.698                        & 0.738                        & 0.754                        & \multicolumn{1}{c|}{0.037}                        & 0.845                        & 0.893                        & 0.767                        & 0.804                        & 0.817                        & 0.048                        \\
\multicolumn{1}{c|}{UGTR}                                         & \multicolumn{1}{c|}{ICCV'21}                                           & 0.785                                                & 0.822                        & 0.685                        & 0.737                        & 0.753                        & \multicolumn{1}{c|}{0.086}                        & 0.818                        & 0.852                        & 0.667                        & 0.712                        & 0.742                        & \multicolumn{1}{c|}{0.035}                        & 0.839                        & 0.874                        & 0.746                        & 0.787                        & 0.807                        & 0.052                        \\
\multicolumn{1}{c|}{ICON}                                         & \multicolumn{1}{c|}{PAMI‘22}                                           & {\color[HTML]{FE0000} 0.840}                         & 0.894                        & {\color[HTML]{FE0000} 0.769} & 0.796                        & {\color[HTML]{FE0000} 0.824} & \multicolumn{1}{c|}{0.058}                        & 0.818                        & 0.904                        & 0.688                        & 0.717                        & 0.756                        & \multicolumn{1}{c|}{0.033}                        & 0.847                        & {\color[HTML]{FE0000} 0.911} & 0.784                        & 0.697                        & 0.817                        & 0.045                        \\
\multicolumn{1}{c|}{TPRNet}                                       & \multicolumn{1}{c|}{TVCJ'22}                                           & 0.807                                                & 0.861                        & 0.725                        & 0.772                        & 0.785                        & \multicolumn{1}{c|}{0.074}                        & 0.817                        & 0.887                        & 0.683                        & 0.724                        & 0.748                        & \multicolumn{1}{c|}{0.036}                        & 0.846                        & 0.898                        & 0.768                        & 0.805                        & 0.820                        & 0.048                        \\ \midrule  
\multicolumn{1}{c|}{SAM}                                       & \multicolumn{1}{c|}{ICCV'23}                                             & 0.684                                                & 0.687                        & 0.606                        & 0.680                        & 0.681                        & \multicolumn{1}{c|}{0.132}                        & 0.783                        & 0.798                        & 0.701                        & 0.756                        & 0.758                        & \multicolumn{1}{c|}{0.050}                        & 0.767                        & 0.776                        & 0.696                        & 0.752                        & 0.754                        & 0.078                        \\ 
\multicolumn{1}{c|}{SAM2}                                       & \multicolumn{1}{c|}{arXiv'24}                                             & 0.444                                                & 0.401                        & 0.184                        & 0.207                        & 0.219                        & \multicolumn{1}{c|}{0.236}                        & 0.549                        & 0.521                        & 0.271                        & 0.291                        & 0.292                        & \multicolumn{1}{c|}{0.134}                        & 0.512                        & 0.482                        & 0.251                       & 0.268                        & 0.269                        & 0.186                        \\  \bottomrule
\end{tabular}}
\label{table1}
\end{table}

\begin{table}[!htbp]
\centering
\caption{Comparison of the number of masks predicted by SAM and SAM2.}
\begin{tabular}{@{}c|c|c|c@{}}
\toprule
     & CAMO (250 Images) & COD10K (2026 Images) & NC4K (4121 Images) \\ \midrule
SAM  & 25472             & 218508               & 347728             \\
SAM2 & 4761              & 33522                & 34044              \\ \bottomrule
\end{tabular}
\label{totalnumber}
\end{table}

\begin{figure*}[!htbp]
    \centering
    \includegraphics[width=0.9\linewidth]{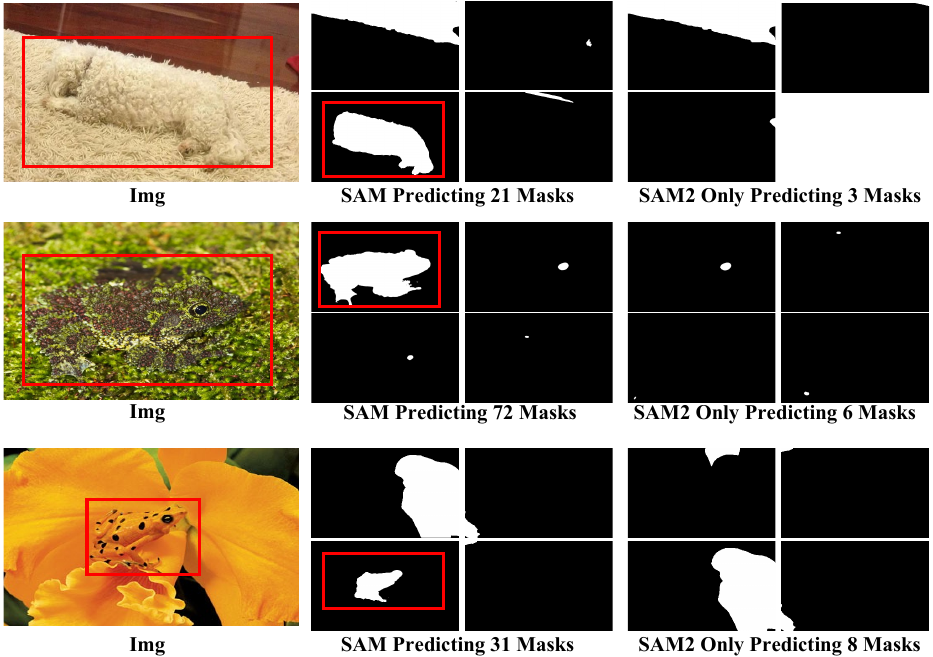}
    \caption{Masks predicted by SAM and SAM2.}
    \label{location}
\end{figure*}

\textbf{Auto Mode.} In auto mode, we evaluate the performance of SAM2 and SAM. The evaluation method is similar to that described in the technical report~\cite{tang2023can}, and the comparison includes 22 CoD methods: SINet~\cite{DBLP:conf/cvpr/FanJSCS020}, C2FNet~\cite{DBLP:conf/ijcai/SunCZZL21}, LSR~\cite{DBLP:conf/cvpr/Lv0DLLBF21}, PFNet~\cite{DBLP:conf/cvpr/MeiJW0WF21}, MGL~\cite{DBLP:conf/cvpr/ZhaiL0C0F21}, JCOD~\cite{DBLP:conf/cvpr/Li0LL0D21}, TANet~\cite{DBLP:conf/aaai/ZhuZZL21}, BGNet~\cite{DBLP:conf/ijcai/SunWCX22}, FDCOD~\cite{DBLP:conf/cvpr/ZhongLTKWD22}, SegMaR~\cite{DBLP:conf/cvpr/0001YL0LL22}, ZoomNet~\cite{DBLP:conf/cvpr/PangZXZL22}, BSANet~\cite{DBLP:conf/aaai/ZhuL0YLCWQ22}, SINetV2~\cite{DBLP:journals/pami/FanJCS22}, FAPNet~\cite{DBLP:journals/tip/ZhouZGYZ22}, the extension version of C2FNet~\cite{DBLP:journals/tcsv/ChenLSJWZ22}, DGNet~\cite{DBLP:journals/ijautcomp/JiFCDLG23}, CubeNet~\cite{DBLP:journals/pr/ZhugeLGCC22}, ERRNet~\cite{DBLP:journals/pr/JiZZF22}, VST~\cite{DBLP:conf/iccv/LiuZW0H21}, UGTR~\cite{DBLP:conf/iccv/0054Z00L0F21}, ICON~\cite{DBLP:journals/pami/ZhugeFLZXS23} and TPRNet~\cite{2022TPRNet}. The performance is shown in Table. \ref{table1}. From the table, it is apparent that in auto mode, SAM2 seems unable to segment potential camouflaged objects. To further illustrate this point, we have compiled statistics on the number of masks predicted by SAM and SAM2 in each dataset, as shown in Table. \ref{totalnumber}. We can see that the number of masks predicted by SAM is six to ten times that predicted by SAM2. As shown in Fig. \ref{location}, for the prediction of masks corresponding to a specific image, SAM2 not only lags significantly behind SAM in terms of quantity, but also in quality.

\clearpage

\section{Conclusion}
This technical report has examined the transition from SAM to SAM2, showcasing both significant advancements and notable limitations. SAM2, an evolution from the foundational SAM, has proven to excel in tasks involving prompt-driven segmentation, where it outperforms SAM with enhanced accuracy and speed, particularly in handling both video and image segmentation. These improvements underscore SAM2’s potential as a versatile tool in the evolving landscape of vision-based models. However, our evaluations, particularly through the lens of camouflaged object detection, highlight a critical area where SAM2 lags behind its predecessor: operating in auto mode without prompts. Here, SAM2’s performance significantly declines, indicating a reliance on prompts that may limit its utility in scenarios demanding autonomous object recognition.
Therefore, we hope continued exploration and refinement of the SAM  family, aiming for advancements that retain the strengths of SAM while overcoming the limitations observed in SAM2. By addressing these challenges, we can push the boundaries of what foundational models can achieve in the realm of computer vision.

\clearpage

{\small
\bibliographystyle{ieee_fullname}
\bibliography{egbib}

\begin{thebibliography}{10}\itemsep=-1pt

\bibitem{DBLP:journals/tcsv/ChenLSJWZ22}
Geng Chen, Si{-}Jie Liu, Yu{-}Jia Sun, Ge{-}Peng Ji, Ya{-}Feng Wu, and Tao
  Zhou.
\newblock Camouflaged object detection via context-aware cross-level fusion.
\newblock {\em {IEEE} Trans. Circuits Syst. Video Technol.}, 32(10):6981--6993,
  2022.

\bibitem{DBLP:journals/corr/abs-2306-15195}
Keqin Chen, Zhao Zhang, Weili Zeng, Richong Zhang, Feng Zhu, and Rui Zhao.
\newblock Shikra: Unleashing multimodal llm's referential dialogue magic.
\newblock {\em CoRR}, abs/2306.15195, 2023.

\bibitem{DBLP:conf/cvpr/ChengXFZHDG22}
Xuelian Cheng, Huan Xiong, Deng{-}Ping Fan, Yiran Zhong, Mehrtash Harandi, Tom
  Drummond, and Zongyuan Ge.
\newblock Implicit motion handling for video camouflaged object detection.
\newblock In {\em {CVPR}}, pages 13854--13863. {IEEE}, 2022.

\bibitem{chowdhery2023palm}
Aakanksha Chowdhery, Sharan Narang, Jacob Devlin, Maarten Bosma, Gaurav Mishra,
  Adam Roberts, Paul Barham, Hyung~Won Chung, Charles Sutton, Sebastian
  Gehrmann, et~al.
\newblock Palm: Scaling language modeling with pathways.
\newblock {\em Journal of Machine Learning Research}, 24(240):1--113, 2023.

\bibitem{DBLP:conf/iccv/FanCLLB17}
Deng{-}Ping Fan, Ming{-}Ming Cheng, Yun Liu, Tao Li, and Ali Borji.
\newblock Structure-measure: {A} new way to evaluate foreground maps.
\newblock In {\em {IEEE} International Conference on Computer Vision, {ICCV}
  2017, Venice, Italy, October 22-29, 2017}, pages 4558--4567. {IEEE} Computer
  Society, 2017.

\bibitem{DBLP:journals/pami/FanJCS22}
Deng{-}Ping Fan, Ge{-}Peng Ji, Ming{-}Ming Cheng, and Ling Shao.
\newblock Concealed object detection.
\newblock {\em {IEEE} Trans. Pattern Anal. Mach. Intell.}, 44(10):6024--6042,
  2022.

\bibitem{DBLP:conf/cvpr/FanJSCS020}
Deng{-}Ping Fan, Ge{-}Peng Ji, Guolei Sun, Ming{-}Ming Cheng, Jianbing Shen,
  and Ling Shao.
\newblock Camouflaged object detection.
\newblock In {\em {Conf. Comput. Vis. Pattern Recog.}}, pages 2774--2784.
  {IEEE}, 2020.

\bibitem{21Fan_HybridLoss}
Deng-Ping Fan, Ge-Peng Ji, Xuebin Qin, and Ming-Ming Cheng.
\newblock Cognitive vision inspired object segmentation metric and loss
  function.
\newblock {\em SCIENTIA SINICA Informationis}, 2021.

\bibitem{hui2024endow}
Wenjun Hui, Zhenfeng Zhu, Shuai Zheng, and Yao Zhao.
\newblock Endow sam with keen eyes: Temporal-spatial prompt learning for video
  camouflaged object detection.
\newblock In {\em Proceedings of the IEEE/CVF Conference on Computer Vision and
  Pattern Recognition}, pages 19058--19067, 2024.

\bibitem{DBLP:journals/ijautcomp/JiFCDLG23}
Ge{-}Peng Ji, Deng{-}Ping Fan, Yu{-}Cheng Chou, Dengxin Dai, Alexander Liniger,
  and Luc~Van Gool.
\newblock Deep gradient learning for efficient camouflaged object detection.
\newblock {\em Int. J. Autom. Comput.}, 20(1):92--108, 2023.

\bibitem{DBLP:journals/pr/JiZZF22}
Ge{-}Peng Ji, Lei Zhu, Mingchen Zhuge, and Keren Fu.
\newblock Fast camouflaged object detection via edge-based reversible
  re-calibration network.
\newblock {\em Pattern Recognit.}, 123:108414, 2022.

\bibitem{DBLP:conf/cvpr/0001YL0LL22}
Qi Jia, Shuilian Yao, Yu Liu, Xin Fan, Risheng Liu, and Zhongxuan Luo.
\newblock Segment, magnify and reiterate: Detecting camouflaged objects the
  hard way.
\newblock In {\em {CVPR}}, pages 4703--4712. {IEEE}, 2022.

\bibitem{kirillov2023segment}
Alexander Kirillov, Eric Mintun, Nikhila Ravi, Hanzi Mao, Chloe Rolland, Laura
  Gustafson, Tete Xiao, Spencer Whitehead, Alexander~C Berg, Wan-Yen Lo, et~al.
\newblock Segment anything.
\newblock {\em arXiv preprint arXiv:2304.02643}, 2023.

\bibitem{DBLP:journals/cviu/LeNNTS19}
Trung{-}Nghia Le, Tam~V. Nguyen, Zhongliang Nie, Minh{-}Triet Tran, and Akihiro
  Sugimoto.
\newblock Anabranch network for camouflaged object segmentation.
\newblock {\em Comput. Vis. Image Underst.}, 184:45--56, 2019.

\bibitem{DBLP:conf/cvpr/Li0LL0D21}
Aixuan Li, Jing Zhang, Yunqiu Lv, Bowen Liu, Tong Zhang, and Yuchao Dai.
\newblock Uncertainty-aware joint salient object and camouflaged object
  detection.
\newblock In {\em {CVPR}}, pages 10071--10081. Computer Vision Foundation /
  {IEEE}, 2021.

\bibitem{DBLP:conf/icml/0001LXH22}
Junnan Li, Dongxu Li, Caiming Xiong, and Steven C.~H. Hoi.
\newblock {BLIP:} bootstrapping language-image pre-training for unified
  vision-language understanding and generation.
\newblock In {\em {Int. Conf. Mach. Learn.}}, volume 162 of {\em Proceedings of
  Machine Learning Research}, pages 12888--12900. {PMLR}, 2022.

\bibitem{DBLP:journals/corr/abs-2304-08485}
Haotian Liu, Chunyuan Li, Qingyang Wu, and Yong~Jae Lee.
\newblock Visual instruction tuning.
\newblock {\em CoRR}, abs/2304.08485, 2023.

\bibitem{DBLP:conf/iccv/LiuZW0H21}
Nian Liu, Ni Zhang, Kaiyuan Wan, Ling Shao, and Junwei Han.
\newblock Visual saliency transformer.
\newblock In {\em {ICCV}}, pages 4702--4712. {IEEE}, 2021.

\bibitem{DBLP:conf/cvpr/Lv0DLLBF21}
Yunqiu Lv, Jing Zhang, Yuchao Dai, Aixuan Li, Bowen Liu, Nick Barnes, and
  Deng{-}Ping Fan.
\newblock Simultaneously localize, segment and rank the camouflaged objects.
\newblock In {\em {Conf. Comput. Vis. Pattern Recog.}}, pages 11591--11601.
  {IEEE}, 2021.

\bibitem{DBLP:conf/cvpr/MargolinZT14}
Ran Margolin, Lihi Zelnik{-}Manor, and Ayellet Tal.
\newblock How to evaluate foreground maps.
\newblock In {\em 2014 {IEEE} Conference on Computer Vision and Pattern
  Recognition, {CVPR} 2014, Columbus, OH, USA, June 23-28, 2014}, pages
  248--255. {IEEE} Computer Society, 2014.

\bibitem{DBLP:conf/cvpr/MeiJW0WF21}
Haiyang Mei, Ge{-}Peng Ji, Ziqi Wei, Xin Yang, Xiaopeng Wei, and Deng{-}Ping
  Fan.
\newblock Camouflaged object segmentation with distraction mining.
\newblock In {\em {Conf. Comput. Vis. Pattern Recog.}}, pages 8772--8781.
  {IEEE}, 2021.

\bibitem{DBLP:journals/corr/abs-2304-07193}
Maxime Oquab, Timoth{\'{e}}e Darcet, and etc~at. Th{\'{e}}o~Moutakanni.
\newblock Dinov2: Learning robust visual features without supervision.
\newblock {\em CoRR}, 2023.

\bibitem{DBLP:conf/cvpr/PangZXZL22}
Youwei Pang, Xiaoqi Zhao, Tian{-}Zhu Xiang, Lihe Zhang, and Huchuan Lu.
\newblock Zoom in and out: {A} mixed-scale triplet network for camouflaged
  object detection.
\newblock In {\em {Conf. Comput. Vis. Pattern Recog.}}, pages 2150--2160.
  {IEEE}, 2022.

\bibitem{DBLP:conf/icml/RadfordKHRGASAM21}
Alec Radford, Jong~Wook Kim, Chris Hallacy, Aditya Ramesh, Gabriel Goh,
  Sandhini Agarwal, Girish Sastry, Amanda Askell, Pamela Mishkin, Jack Clark,
  Gretchen Krueger, and Ilya Sutskever.
\newblock Learning transferable visual models from natural language
  supervision.
\newblock In {\em {{Int. Conf. Mach. Learn.}}}, volume 139 of {\em Proceedings
  of Machine Learning Research}, pages 8748--8763. {PMLR}, 2021.

\bibitem{ravi2024sam2}
Nikhila Ravi, Valentin Gabeur, Yuan-Ting Hu, Ronghang Hu, Chaitanya Ryali,
  Tengyu Ma, Haitham Khedr, Roman R{\"a}dle, Chloe Rolland, Laura Gustafson,
  Eric Mintun, Junting Pan, Kalyan~Vasudev Alwala, Nicolas Carion, Chao-Yuan
  Wu, Ross Girshick, Piotr Doll{\'a}r, and Christoph Feichtenhofer.
\newblock Sam 2: Segment anything in images and videos.
\newblock {\em arXiv preprint}, 2024.

\bibitem{DBLP:conf/ijcai/SunCZZL21}
Yujia Sun, Geng Chen, Tao Zhou, Yi Zhang, and Nian Liu.
\newblock Context-aware cross-level fusion network for camouflaged object
  detection.
\newblock In {\em {Int. Joint Conf. Artif. Intell}}, pages 1025--1031.
  ijcai.org, 2021.

\bibitem{DBLP:conf/ijcai/SunWCX22}
Yujia Sun, Shuo Wang, Chenglizhao Chen, and Tian{-}Zhu Xiang.
\newblock Boundary-guided camouflaged object detection.
\newblock In {\em {Int. Joint Conf. Artif. Intell}}, pages 1335--1341.
  ijcai.org, 2022.

\bibitem{tang2024chain}
Lv Tang, Peng-Tao Jiang, Zhihao Shen, Hao Zhang, Jinwei Chen, and Bo Li.
\newblock Chain of visual perception: Harnessing multimodal large language
  models for zero-shot camouflaged object detection.
\newblock In {\em ACM Multimedia 2024}, 2024.

\bibitem{tang2023can}
Lv Tang, Haoke Xiao, and Bo Li.
\newblock Can sam segment anything? when sam meets camouflaged object
  detection.
\newblock {\em arXiv preprint arXiv:2304.04709}, 2023.

\bibitem{touvron2023llama}
Hugo Touvron, Thibaut Lavril, Gautier Izacard, Xavier Martinet, Marie-Anne
  Lachaux, Timoth{\'e}e Lacroix, Baptiste Rozi{\`e}re, Naman Goyal, Eric
  Hambro, Faisal Azhar, et~al.
\newblock Llama: Open and efficient foundation language models.
\newblock {\em arXiv preprint arXiv:2302.13971}, 2023.

\bibitem{DBLP:conf/iccv/0054Z00L0F21}
Fan Yang, Qiang Zhai, Xin Li, Rui Huang, Ao Luo, Hong Cheng, and Deng{-}Ping
  Fan.
\newblock Uncertainty-guided transformer reasoning for camouflaged object
  detection.
\newblock In {\em {Int. Conf. Comput. Vis.}}, pages 4126--4135. {IEEE}, 2021.

\bibitem{DBLP:conf/cvpr/ZhaiL0C0F21}
Qiang Zhai, Xin Li, Fan Yang, Chenglizhao Chen, Hong Cheng, and Deng{-}Ping
  Fan.
\newblock Mutual graph learning for camouflaged object detection.
\newblock In {\em {Conf. Comput. Vis. Pattern Recog.}}, pages 12997--13007.
  {IEEE}, 2021.

\bibitem{2022TPRNet}
Qiao Zhang, Yanliang Ge, Cong Zhang, and Hongbo Bi.
\newblock Tprnet: camouflaged object detection via transformer-induced
  progressive refinement network.
\newblock {\em The Visual Computer}, pages 1--15, 2022.

\bibitem{zhang2022opt}
Susan Zhang, Stephen Roller, Naman Goyal, Mikel Artetxe, Moya Chen, Shuohui
  Chen, Christopher Dewan, Mona Diab, Xian Li, Xi~Victoria Lin, et~al.
\newblock Opt: Open pre-trained transformer language models.
\newblock {\em arXiv preprint arXiv:2205.01068}, 2022.

\bibitem{DBLP:conf/cvpr/ZhongLTKWD22}
Yijie Zhong, Bo Li, Lv Tang, Senyun Kuang, Shuang Wu, and Shouhong Ding.
\newblock Detecting camouflaged object in frequency domain.
\newblock In {\em {Conf. Comput. Vis. Pattern Recog.}}, pages 4494--4503.
  {IEEE}, 2022.

\bibitem{DBLP:journals/tip/ZhouZGYZ22}
Tao Zhou, Yi Zhou, Chen Gong, Jian Yang, and Yu Zhang.
\newblock Feature aggregation and propagation network for camouflaged object
  detection.
\newblock {\em {IEEE} Trans. Image Process.}, 31:7036--7047, 2022.

\bibitem{DBLP:conf/aaai/ZhuL0YLCWQ22}
Hongwei Zhu, Peng Li, Haoran Xie, Xuefeng Yan, Dong Liang, Dapeng Chen,
  Mingqiang Wei, and Jing Qin.
\newblock I can find you! boundary-guided separated attention network for
  camouflaged object detection.
\newblock In {\em {AAAI}}, pages 3608--3616. {AAAI} Press, 2022.

\bibitem{DBLP:conf/aaai/ZhuZZL21}
Jinchao Zhu, Xiaoyu Zhang, Shuo Zhang, and Junnan Liu.
\newblock Inferring camouflaged objects by texture-aware interactive guidance
  network.
\newblock In {\em {AAAI}}, pages 3599--3607. {AAAI} Press, 2021.

\bibitem{DBLP:journals/pami/ZhugeFLZXS23}
Mingchen Zhuge, Deng{-}Ping Fan, Nian Liu, Dingwen Zhang, Dong Xu, and Ling
  Shao.
\newblock Salient object detection via integrity learning.
\newblock {\em {IEEE} Trans. Pattern Anal. Mach. Intell.}, 45(3):3738--3752,
  2023.

\bibitem{DBLP:journals/pr/ZhugeLGCC22}
Mingchen Zhuge, Xiankai Lu, Yiyou Guo, Zhihua Cai, and Shuhan Chen.
\newblock Cubenet: X-shape connection for camouflaged object detection.
\newblock {\em Pattern Recognit.}, 127:108644, 2022.

\end{thebibliography}
}

\end{document}